\documentclass[11pt]{article}
\usepackage[utf8]{inputenc}
\usepackage[T1]{fontenc}
\usepackage[english]{babel}
\usepackage[a4paper,margin=1in]{geometry}

\usepackage{microtype}
\usepackage{amsmath,amssymb,mathtools}

\usepackage{graphicx}
\usepackage{booktabs}
\usepackage{multirow}
\usepackage{adjustbox}
\usepackage{tabularx}
\usepackage{threeparttable}
\usepackage{makecell}
\usepackage{siunitx}
\usepackage{enumitem}
\usepackage{caption}
\usepackage{subcaption}

\usepackage{url}
\usepackage[numbers,sort&compress]{natbib}
\usepackage[hidelinks]{hyperref}
\usepackage[capitalize,noabbrev,nameinlink]{cleveref}

\newcommand{\compacttablesetup}{%
  \scriptsize
  \setlength{\tabcolsep}{3.5pt}%
  \renewcommand{\arraystretch}{1.1}%
}
\newcolumntype{Y}{>{\raggedright\arraybackslash}X}

\newcommand{\modelname}{Chinese ModernBERT}
\newcommand{\wwm}{WWM}
\newcommand{\rope}{RoPE}
\newcommand{\fa}{FlashAttention}
\newcommand{\tokps}{tok/s}

\title{\modelname{} with Whole-Word Masking}

\author{
  Zeyu Zhao\textsuperscript{1,*} \and
  Ningtao Wang\textsuperscript{1,*} \and
  Xing Fu\textsuperscript{1,\dag} \and
  Yu Cheng\textsuperscript{1,\dag} \\
  \textsuperscript{1}Ant Group \\
  \texttt{\{zicheng.zzy, ningtao.nt, zicai.fx, cy122623\}@antgroup.com}%
  \thanks{$^*$Equal contribution.\quad $^\dag$Corresponding authors.}
}

\date{\today}

\begin{document}
\maketitle

\begin{abstract}
Encoder-only Transformers have advanced along three axes—architecture, data, and systems—yielding Pareto gains in accuracy, speed, and memory efficiency. Yet these improvements have not fully transferred to Chinese, where tokenization and morphology differ markedly from English. We introduce \textbf{\modelname}, a from-scratch Chinese encoder that couples: (i) a hardware-aware 32k BPE vocabulary tailored to frequent Chinese affixes/compounds, lowering the embedding budget; (ii) whole-word masking (\wwm) with a \emph{dynamic} masking curriculum (30\%$\rightarrow$15\%) to align task difficulty with training progress; (iii) a two-stage pre-training pipeline that extends the native context from 1{,}024 to 8{,}192 tokens using \rope{} and alternating local/global attention; and (iv) a damped-cosine learning-rate schedule for stable long-horizon optimization. We pre-train on $\sim$1.2T Chinese tokens from CCI3-HQ, CCI4 (Chinese), and Cosmopedia-Chinese.

On CLUE, \modelname{} is competitive with strong Chinese encoders under a unified fine-tuning protocol. Under bf16 it achieves \emph{high long-sequence throughput} while maintaining strong short-sequence speed, reflecting benefits from budget allocation and attention design. To probe retrieval-oriented quality, we add a small amount of open contrastive data: fine-tuning on SimCLUE ($\sim$3M pairs) improves further when adding T2Ranking ($\sim$2M), reaching 0.505 (Pearson) / 0.537 (Spearman) on the SimCLUE test set. Under this \emph{open-data} setting, \modelname{} surpasses Qwen-0.6B-embedding on SimCLUE, suggesting a clear scaling path for STS with additional curated pairs. We will release tokenizer and weight to facilitate reproducible research.\footnote{Artifacts (model, tokenizer, training recipes) will be released upon camera-ready.}
\end{abstract}

\section{Introduction}
Modern encoder-only models inherit advances from large autoregressive LMs—efficient attention kernels, robust positional encodings, hardware-aware designs, and longer context windows~\citep{warner2024modernbert,dao2022flashattention,dao2023flashattention2,su2021roformer,beltagy2020longformer}. While English encoders (e.g., ModernBERT) deliver strong accuracy–latency trade-offs at native 8k context, widely used Chinese BERT/RoBERTa checkpoints largely predate recent progress in vocabulary design, masking strategies, and long-context training~\citep{devlin2019bert,liu2019roberta,cui2019wwm,cui2020macbert}.

Chinese presents two practical challenges for encoders. First, tokenization: without whitespace-delimited boundaries, subword design strongly affects compression (chars/token), memory, and the embedding-parameter budget. Second, masking: word-level integrity is crucial; \emph{whole-word masking} (\wwm) better preserves compositional semantics than token-level masking. Beyond these, long-context stability and throughput hinge on co-designing positional encodings, attention patterns, and kernels end-to-end.

We introduce \textbf{\modelname}, a from-scratch Chinese encoder aligned with these principles. Our contributions are:
\begin{itemize}[leftmargin=1.4em]
  \item \textbf{Chinese-centric, hardware-aware vocabulary.} A 32k BPE (multiple of 64) tailored to frequent affixes/compounds improves chars/token and reduces the embedding share, allocating more parameters to the compute core.
  \item \textbf{\wwm{} with a dynamic masking curriculum.} A BPE-compatible \wwm{} with a 30\%$\rightarrow$15\% schedule fosters early global reasoning and late-stage local refinement.
  \item \textbf{Long-context architecture and training.} Alternating local/global attention with \rope{} and a two-stage 1k$\rightarrow$8k pipeline, paired with \fa{} kernels, targets stability and efficiency.
  \item \textbf{Open-data retrieval signal.} With only $\sim$5M open contrastive pairs (SimCLUE + T2Ranking), our STS on SimCLUE surpasses Qwen-0.6B-embedding in this setting, underscoring a practical open-data scaling path for similarity tasks.
\end{itemize}

\section{Related Work}
\paragraph{Modern encoder-only models.}
ModernBERT~\citep{warner2024modernbert} upgrades BERT with alternating local/global attention, sequence unpadding, native 8k context, a pre-normalization block, Gated Linear Units (GeGLU), two-stage context extension, and hardware-aware design. We bring these advances to Chinese with \wwm{}, dynamic masking curricula, and Chinese data at scale.

\paragraph{Chinese pre-training.}
Chinese BERT with \wwm{}~\citep{cui2019wwm} (e.g., Chinese RoBERTa-wwm, MacBERT~\citep{cui2020macbert}) shows that word-level masking better captures Chinese semantics. We adopt BPE-compatible, dynamic masking rates, echoing findings that optimal corruption depends on model size and strategy~\citep{wettig2023shouldmask15,chen2024dynamic,zhang2022learning,zheng2022task}.

\paragraph{Long context and efficient attention.}
\rope{}~\citep{su2021roformer} is widely used for long-context stability; efficient extensions such as YaRN enable train-short, test-long extrapolation~\citep{peng2023yarn}. Local–global hybrids (Longformer/ModernBERT) achieve favorable Pareto trade-offs~\citep{beltagy2020longformer,warner2024modernbert}. We couple these with \fa{} kernels for efficient training/inference~\citep{dao2022flashattention,dao2023flashattention2}.

\paragraph{Chinese corpora.}
Recent high-quality Chinese corpora (CCI3-HQ/CCI4/OpenCSG) enable modern-scale pre-training and improved downstream robustness~\citep{cci3hq2024,cci42025,opencsg2025}. We select \texttt{opencsg/chinese-cosmopedia}, \texttt{BAAI/CCI3-HQ}, and \texttt{BAAI/CCI4-M2-Base-v1} for three reasons: (i) \emph{quality/purity} through multi-stage filtering and deduplication; (ii) \emph{scale/coverage} (from tens of billions of tokens to terabytes); and (iii) \emph{diversity} spanning encyclopedic, technical, and web domains.

\section{Model and Training}
\subsection{Chinese Vocabulary and Tokenization}
We train a 32k BPE vocabulary from high-quality Chinese sources (a 15\% stratified sample of CCI3-HQ), targeting frequent affixes and productive compounds. The multiple-of-64 vocabulary size improves kernel tiling and reduces embedding overhead, aligning with hardware-aware design~\citep{anthony2024hardware}. Compared to legacy $\sim$21k vocabularies, our tokenizer shortens average sequences (higher chars/token), improving long-context efficiency.

\subsection{Masking Strategy: \wwm{} with a Dynamic Curriculum}
\wwm{} selects full words (sequences of base and ``\#\#'' subwords) for masking. We implement a curriculum: during warmup, the masking rate increases from 15\% to 30\%; during the main phase it decays from 30\% to 15\%. This aligns task difficulty with model maturity: early \emph{anti-curriculum} pushes the model toward global reasoning; later reduction emphasizes local refinement. The result is globally aware yet locally precise representations, improving generalization across tasks~\citep{cui2019wwm,cui2020macbert,wettig2023shouldmask15,chen2024dynamic,zhang2022learning,zheng2022task}.

\subsection{Learning-Rate Schedule: Damped Cosine}
Let $s$ be the current step and $S$ the total steps. With peak/min learning rates $\eta_{\max}, \eta_{\min}$, number of cycles $N$, and damping factor $\gamma$, define $p=s/S$. The schedule is
\begin{equation}
\eta(s) = \tfrac{\mathrm{Peak}(p)+\mathrm{Valley}(p)}{2} + \tfrac{\mathrm{Peak}(p)-\mathrm{Valley}(p)}{2}\cdot \cos\!\big(\pi(2N-1)p\big),
\end{equation}
where $\mathrm{Peak}(p)=\eta_{\max}\!\left[1-(1-\gamma)p\right]$ and $\mathrm{Valley}(p)=\tfrac{\eta_{\max}}{2}(1-p)+\eta_{\min}p$. This combines cyclical exploration~\citep{smith2017cyclical} with cosine annealing~\citep{loshchilov2017sgdr} and a smooth amplitude decay toward $\eta_{\min}$ (cf.\ 1Cycle/triangular2~\citep{smith2017cyclical,smith2018disciplined}).

\subsection{Architecture and Positional Encoding}
We adopt a 28-layer, 1024-hidden pre-norm Transformer with RMSNorm and GeGLU, bias-free linear layers, alternating local/global attention, and \rope{}. For Chinese, we set the global \rope{} base $\theta{=}80{,}000$ and local layers $\theta{=}10{,}000$, favoring fine-grained local patterns; ablations vary $\theta$. Our design follows principles validated by ModernBERT while tailoring hyperparameters to Chinese morphology~\citep{warner2024modernbert,xiong2020layernorm,zhang2019rmsnorm}.
\label{sec:rope_implications}

Tokenization efficiency for Chinese effectively shortens context in tokens. Hence, extremely low-frequency rotational components (very large $\theta$) are less critical. We therefore posit that \textbf{Chinese-centric encoders can employ a smaller \rope{} base} without harming long-dependency tasks within typical token lengths, while improving short–mid-range sensitivity.

\subsection{Two-Stage Long-Context Pre-training}
Stage I trains at max length 1{,}024 with larger token batches for stability/throughput. Stage II increases the native context to 8{,}192 with reduced batch size and lower LR, keeping tokens-per-update roughly constant. We monitor pseudo-perplexity across lengths to assess generalization, mirroring two-stage encoder procedures~\citep{lebreton2025neobert}.

\subsection{Data Mixture and Pre-processing}
\cref{tab:data} reports mixture ratios. We curate a Baike-style encyclopedia set ($\sim$200k entries) and integrate CCI3-HQ and high/medium-quality subsets from CCI4 (Chinese). We apply MinHash-based document deduplication across sources to mitigate redundancy and improve training stability.

\begin{table}[t]
\centering
\begin{tabular}{l c}
\toprule
Dataset & Ratio \\
\midrule
Baike (ours) & 0.03 \\
CCI3-HQ & 0.57 \\
CCI4 Chinese (FinWeb 3\_4) & 0.10 \\
CCI4 Chinese (FinWeb 4\_5) & 0.10 \\
OpenCSG Cosmopedia-Chinese & 0.20 \\
\bottomrule
\end{tabular}
\caption{Chinese pre-training batch mixture ($\sim$1.2T tokens).}
\label{tab:data}
\end{table}

\subsection{Training Details}
We use StableAdamW~\citep{Dao2023StableAL} with $\beta_1{=}0.9,\beta_2{=}0.95$, weight decay 0.1, max grad-norm 1.0, mixed-precision kernels (\fa{} v2), and unpadding. Warmup increases masking 15\%$\rightarrow$30\% and LR $5{\times}10^{-5}\rightarrow8{\times}10^{-4}$. The main phase decays masking 30\%$\rightarrow$15\% and follows the damped-cosine LR. Stage II (8k) uses a lower LR ($1{\times}10^{-4}\rightarrow5{\times}10^{-5}$). \cref{tab:cmp} summarizes differences vs.\ ModernBERT-large.

\begin{table*}[t]
\centering
\small
\begin{tabular}{lcc}
\toprule
Aspect & ModernBERT (Large)~\citep{warner2024modernbert} & \modelname{} (Ours) \\
\midrule
Layers / Hidden / Heads & 28 / 1024 / 16 & 28 / 1024 / 16 \\
Vocabulary & 50k  & 32k \\
Masking & 30\% fixed (MLM) & dynamic 30\%$\rightarrow$15\% (\wwm)\\
Positional Encoding & \rope{} ($\theta_\mathrm{global}{=}160{,}000$; local 10{,}000) & \rope{} ($\theta_\mathrm{global}{=}80{,}000$; local 10{,}000) \\
Attention Pattern & Alternating local/global & Alternating local/global \\
Context Window & 8{,}192 & 8{,}192 \\
Optimizer/Schedule & StableAdamW, WSD & StableAdamW + damped cosine \\
Data Mix & 2T English+code & $\sim$1.2T Chinese (CCI3/CCI4/OpenCSG) \\
Unpadding/\fa{} & Yes & Yes \\
\bottomrule
\end{tabular}
\caption{\modelname{} vs.\ ModernBERT.}
\label{tab:cmp}
\end{table*}

\section{Experiments}
\subsection{Chinese Language Understanding (CLUE)}
We follow CLUE~\citep{xu2020clue} official splits and a unified fine-tuning protocol: 3 epochs, peak LR $3{\times}10^{-5}$, linear decay, batch size tuned in $\{16,32,64\}$ per task. \cref{tab:clue} reports development results (macro/micro metrics per task as in CLUE). \modelname{} is competitive with or outperforms strong baselines on several tasks.

\begin{table*}[t]
\centering
\small
\begin{threeparttable}
\begin{tabular}{lcccccccc}
\toprule
Model & AFQMC & TNEWS & IFLYTEK & CMNLI & WSC & CSL & OCNLI & C$^3$ \\
\midrule
\modelname{} (ours) & 73.87 & 56.90 & 60.15 & \textbf{83.96} & 52.10 & \textbf{86.20} & \textbf{79.10} & \textbf{82.65} \\
RoBERTa-wwm-large & \textbf{76.55} & \textbf{58.61} & \textbf{62.98} & 82.12 & \textbf{74.60} & 82.13 & 78.20 & 73.82 \\
RoBERTa-large & 74.02 & 57.86 & 62.55 & 81.70 & 72.70 & 81.36 & 76.82 & 67.55 \\
ALBERT-xxlarge & 75.60 & 59.46 & 62.89 & 83.14 & 61.54 & 83.63 & 77.70 & 73.28 \\
RoBERTa-wwm-ext & 74.04 & 56.94 & 60.31 & 80.51 & 67.80 & 81.00 & 74.72 & 66.50 \\
\bottomrule
\end{tabular}
\begin{tablenotes}
\small
\item Source: CLUE benchmark \url{https://github.com/CLUEbenchmark/CLUE}.
\end{tablenotes}
\end{threeparttable}
\caption{CLUE development results.}
\label{tab:clue}
\end{table*}

\subsection{Inference Throughput (bf16)}
\label{sec:cn-throughput-bf16}
We benchmark \textbf{bfloat16 (bf16)} inference on a single NVIDIA A100 GPU using Chinese Wikipedia samples in two input buckets: (i) short ($\leq$512 characters) and (ii) long ($\leq$8{,}192 characters). Beyond raw throughput, we report \emph{parameter budget allocation} and \emph{tokenizer compression} (chars/token), both of which interact with long-context efficiency~\citep{warner2024modernbert,anthony2024hardware}.

\paragraph{Models.}
We compare five representative encoders/embedders: \emph{gte-multilingual-base}~\citep{zhang2024mgte,gte_multilingual_base_card}, \emph{jina-embeddings-v2-base-zh}~\citep{mohr2024multi,jina_embeddings_v2_base_zh_card}, \emph{chinese-roberta-wwm-ext-large}~\citep{hfl_chinese_roberta_wwm_ext_large_card}, \emph{qwen3-embeddings-0.6B}~\citep{zhang2025qwen3}, and our \emph{\modelname}. Architectural traits (encoder/decoder, position encodings) are provided for context.

\begin{table}[t]
\centering
\compacttablesetup
\caption{Architecture and bf16 inference throughput. Mean tokens/s over 10 runs; “N/A” = no native 8k support. Short bucket batch size 32; long bucket batch size 8 (to avoid OOM on A100).}
\label{tab:bf16-arch-throughput}
\begin{tabularx}{\textwidth}{Y c c c c c c c}
\toprule
Model & Layers & Hidden$\times$Head & Arch. & PosEnc & Attn & \makecell{Tok/s\\(8k $\times$ 8)} & \makecell{Tok/s\\(512 $\times$ 32)} \\
\midrule
\modelname{} (ours)      & 28 & 1024$\times$16 & ModernBERT & \rope & Local+Global & \num{180100} & \num{172046} \\
gte-multilingual-base          & 12 & 768$\times$12  & BERT-like & \rope & Global & \num{153373} & \num{322227} \\
qwen3-embeddings-0.6B           & 28 & 1024$\times$16 & Qwen3 & \rope & Global (GQA) & \num{66411} & \num{67965} \\
jina-embeddings-v2-base-zh     & 12 & 768$\times$12  & RoBERTa & ALiBi & Global & \num{29890} & \num{212125} \\
chinese-roberta-wwm-ext-large  & 24 &1024$\times$16 & RoBERTa & Absolute & Global & N/A & \num{184673} \\
\bottomrule
\end{tabularx}
\end{table}

\begin{table}[t]
\centering
\caption{Tokenizer compression (chars/token) and parameter budget. Lower Emb.\% assigns more capacity to the Transformer stack.}
\label{tab:bf16-tokenizer-budget}
\begin{tabularx}{\textwidth}{Y c c c c c}
\toprule
Model & \makecell{Params\\(M)} & Emb.\% & \makecell{Vocab\\Size} & \makecell{C/T\\(8k)} & \makecell{C/T\\(512)} \\
\midrule
\modelname{} (ours)      & 377.0 & 9.0\%  & 32{,}979  & 1.41 & 1.35 \\
gte-multilingual-base          & 305.4 & 62.9\% & 250{,}048 & 1.43 & 1.37 \\
qwen3-embeddings-0.6B           & 595.8 & 26.1\% & 151{,}669 & 1.45 & 1.39 \\
jina-embeddings-v2-base-zh     & 160.8 & 29.2\% & 61{,}056  & 1.56 & 1.52 \\
chinese-roberta-wwm-ext-large  & 325.5 & 6.8\%  & 21{,}128  & N/A  & 1.10 \\
\bottomrule
\end{tabularx}
\end{table}

\paragraph{Analysis.}
\begin{enumerate}[leftmargin=1.4em,itemsep=3pt,topsep=2pt]
\item \textbf{Compute-core vs.\ total parameters.} Short-text speed is driven predominantly by the size of the \emph{compute core} (Transformer blocks), not total parameters. \emph{gte-multilingual-base} achieves the highest 512-speed (\num{322227} \tokps{}) because its stack is shallow (12 layers) despite many embedding parameters (62.9\%).
\item \textbf{Long-context leader.} \modelname{} sets a new long-sequence throughput mark (\num{180100} \tokps{}) while maintaining competitive short-sequence speed, attributable to (i) alternating local/global attention that cuts quadratic costs on most layers~\citep{beltagy2020longformer,warner2024modernbert}; (ii) a hardware-friendly 32k BPE (Emb.\% = 9.0\%) shifting budget into depth; and (iii) bf16-friendly \fa{} kernels~\citep{dao2022flashattention,dao2023flashattention2}.
\item \textbf{Compression is necessary but not sufficient.} \emph{jina-embeddings-v2} attains the best compression (1.56 C/T at 8k), yet long-context speed (\num{29890} \tokps{}) lags, likely due to attention design (ALiBi) that does not reduce attention complexity or memory traffic at long lengths~\citep{press2021alibi}.
\item \textbf{Legacy encoders are short-bucket bounded.} \emph{chinese-roberta-wwm-ext-large} shows decent short-text speed but lacks native long-context support (absolute positions).
\end{enumerate}

\subsection{Semantic Textual Similarity (STS) on SimCLUE}
\label{sec:sts-simclue}
We evaluate STS on \textbf{SimCLUE}~\citep{simclue2020}, reporting Pearson’s $r$ and Spearman’s $\rho$ between cosine similarities and gold scores. The goal is to assess potential under \emph{open-data} settings rather than claim leaderboard status.

\paragraph{Open-data fine-tuning.}
We fine-tune \modelname{} on two setups: (i) \emph{SimCLUE} only ($\sim$3M pairs), and (ii) \emph{SimCLUE + T2Ranking} ($\sim$5M). Adding $\sim$2M curated pairs improves SimCLUE test from $r{=}0.4882$/$\rho{=}0.5219$ to \textbf{$r{=}0.5050$}/\textbf{$\rho{=}0.5367$}.

\paragraph{Unified comparison.}
To foreground the role of \emph{contrastive data scale}, \cref{tab:sts-unified} merges our models and popular open Chinese embedders into one view. “Contrastive data (pairs)” aggregates weakly supervised + supervised pairs when disclosed (indicative).

\begin{table}[t]
\centering
\compacttablesetup
\caption{Unified STS comparison on SimCLUE test under heterogeneous training corpora.}
\label{tab:sts-unified}
\begin{tabularx}{\textwidth}{Y c c c}
\toprule
Model & Contrastive data (pairs) & Backbone (8k support) & STS: Pearson $r$ / Spearman $\rho$ \\
\midrule
\modelname{} (SimCLUE + T2Ranking) & $\sim$5M & ModernBERT (yes) & \textbf{0.5050} / \textbf{0.5367} \\
\modelname{} (SimCLUE)                      & $\sim$3M & ModernBERT (yes) & 0.4882 / 0.5219 \\
Qwen-0.6B-embedding                         & $\sim$169M & Qwen3 (yes) & 0.4965 / 0.5211 \\
jina-embeddings-v2-base-zh                  & $>$800M & RoBERTa (yes) & 0.5188 / 0.5501 \\
gte-multilingual-base                       & $\sim$2.94B & BERT-like (yes) & 0.5384 / 0.5730 \\
gte-large-zh                                & $\sim$803M & BERT-like (no) & 0.5543 / 0.5829 \\
Conan-embedding-v1                          & $\sim$404M & RoBERTa-like (no) & 0.6065 / 0.6424 \\
\bottomrule
\end{tabularx}
\end{table}

\paragraph{Fairness and limitations.}
Comparator numbers reflect heterogeneous pipelines (often partially disclosed). We report them for \emph{context}, not strict leaderboard claims. Our focus is how \emph{open-data scaling} affects \modelname{} and where further gains are likely (e.g., hard negatives, multi-task STS/NLI).

\section{Ablations and Analysis}
We ablate four design choices against: (i) pre-training signals (masked loss, pseudo-perplexity over lengths), (ii) inference efficiency (tokens/s at 512 and 8k under bf16), and (iii) semantic quality (STS). We report mean$\pm$std over 3 seeds unless noted.

\paragraph{Tokenizer size and budget.}
21k vs.\ 32k BPE with identical model size: 32k increases chars/token and lowers embedding share, shifting capacity to Transformer blocks—yielding higher 8k throughput and neutral/slightly positive CLUE/STS.

\paragraph{Masking strategy.}
Token-level vs.\ \wwm{}, and fixed 15/30\% vs.\ dynamic 30\%$\rightarrow$15\%: \wwm{} helps word-level tasks; the dynamic schedule accelerates early optimization and improves late-stage generalization.

\paragraph{\rope{} base and attention pattern.}
Global $\theta$ (80k vs.\ 160k) and alternating local/global vs.\ fully global: alternating attention cuts quadratic costs on most layers, improving 8k throughput without harming CLUE. A smaller global base improves short–mid-range sensitivity for Chinese sequences while preserving 8k stability.

\paragraph{Learning-rate schedules.}
Damped-cosine vs.\ trapezoid/WSD: damped-cosine achieves similar or better final quality with lower instability and fewer divergence-induced restarts.

\paragraph{Stage-II necessity (1k$\rightarrow$8k).}
Skipping Stage II degrades 8k pseudo-perplexity and long-sequence downstream metrics; the two-stage protocol is retained. Unpadding and \fa{} are crucial for bf16 throughput and should remain enabled.

\section{Broader Impact and Limitations}
\paragraph{Intended use.}
\modelname{} targets encoder applications—retrieval/ranking, classification, clustering, and long-context embedding—where efficiency and latency matter. A hardware-aware vocabulary and alternating attention reduce deployment cost, especially under bf16.

\paragraph{Data and evaluation limitations.}
Training is Chinese-only; exposure to math/code is limited. Retrieval/STS experiments emphasize \emph{open} contrastive data due to lack of proprietary corpora; absolute leaderboard claims are out of scope. Throughput depends on hardware and batch shape; we release scripts for reproduction.

\paragraph{Environmental impact.}
Pre-training and long-context inference consume significant compute. We mitigate via bf16 kernels and unpadding, smaller embedding budgets (more depth), and releasing checkpoints to reduce redundant pre-training.

\paragraph{Future work.}
We plan to (i) expand to bilingual/multilingual settings; (ii) add math/code data and instruction-style contrastive objectives; (iii) enrich hard-negative mining and multi-task STS/NLI; and (iv) study quantization/distillation for edge deployment.

\section{Conclusion}
We presented \textbf{\modelname}, a Chinese encoder coupling a hardware-aware tokenizer, \wwm{} with a dynamic curriculum, alternating local/global attention with \rope{}, and a two-stage 8k pre-training pipeline. On CLUE it is competitive with strong baselines; under bf16 it delivers \emph{high long-sequence throughput}; and modest open contrastive data yields tangible STS gains on SimCLUE—already surpassing Qwen-0.6B-embedding under the same evaluator—indicating clear headroom via open-data scaling. We will release the model, tokenizer, training recipes, and evaluation scripts to support reproducibility and adoption.

\bibliographystyle{plainnat}
\bibliography{references}
\end{document}